# Structured Cognitive Loop for Behavioral Intelligence in Large Language Model Agents

*Extended Revision: From Behavioral Architecture to Epistemic Accountability*

**Myung Ho Kim**
Professor, JEI University
enkiluv@gmail.com
ORCID: 0009-0001-3709-7622

## Abstract

The central challenge for AI agents is not performance but accountability. An agent that produces correct outputs through an opaque process provides no basis for verifying why a decision was permitted, attributing errors to identifiable causes, or delegating authority with justified confidence. Existing LLM-based agent frameworks largely treat performance improvement as the primary objective, intertwining cognition, control, and memory in a single prompt in ways that obscure the justificatory basis of any action taken. The Structured Cognitive Loop (SCL) is introduced as an architectural alternative that separates these functions. In SCL, the language model is dedicated to cognition only, execution is governed by a lightweight controller that mediates every proposal before any action is taken, and memory is maintained externally as a structured store of verified facts. This design makes the justificatory basis of each decision structurally explicit rather than implicitly distributed across prompt tokens. We evaluate SCL against prompt-based baselines including ReAct and common LangChain agents across three scenarios: temperature-based travel planning, email drafting with conditional send, and constraint-guided image generation. All systems share the same base model and tools under matched decoding settings. Across 360 episodes, SCL shows modest but consistent improvements. Task success averages 86.3 percent, compared with 70.5–76.8 percent for prompt-based baselines. Goal fidelity is higher, redundant calls are fewer, intermediate states are reused more reliably, and unsupported assertions are reduced. Ablations show that control and external memory each contribute independently, and decoding sweeps confirm stability of the effects. These results suggest that architectural separation improves reliability and traceability without requiring larger models or heavier prompts. The present study focuses on controlled settings; broader validation across model families and task domains remain open. Three subsequent lines of work extend the SCL framework toward a more complete account of accountable artificial agency: context-aware Human-in-the-Loop, which integrates human judgment as a first-class cognitive process; Pool-Gated Retrieval (PGR), which grounds generation in warranted evidence and registers confirmed absences as first-class facts; and the Horizon-Warrant-Commitment (H-W-C) framework, which provides the philosophical foundation unifying both. The most recent development of the framework moves the loop itself from interpretation toward compilation: instructions are compiled into a program that is validated before execution, so that execution authority derives from a verified program rather than from per-cycle model interpretation, and the language model is confined to residual semantic judgment. For the kind of accountability it pursues, SCL adopts neither multi-agent consensus as a proxy for truth nor the imitation of human cognition as a design goal: it institutionalizes cognition rather than imitating it, and is in this sense an accountability structure rather than a more human-like agent. Together, SCL, Pool-Gated Retrieval, and H-W-C constitute three mutually reinforcing axes of an epistemic accountability architecture—one in which agents are not merely capable but structurally answerable for every decision they make.

## 1 Introduction: Toward Structured Cognition for Behavioral Intelligence

Large language models (LLMs) have rapidly advanced the capabilities of artificial intelligence in natural language understanding and generation, excelling in tasks such as summarization, translation, and question answering (OpenAI, 2023). These achievements have fueled the development of LLM-based agents that can plan trips, manage schedules, or coordinate workflows through natural language interaction (Shinn et al., 2023; Patel et al., 2023). Despite such progress, there is growing recognition that beyond linguistic performance, agents must also demonstrate **behavioral intelligence** - the ability to reason, remember, and act coherently across multi-step tasks in ways that are interpretable, reliable, and adaptable (Griffiths et al., 2019; Lake et al., 2017).

Recent implementations highlight both the promise and fragility of this pursuit. Systems such as ReAct (Yao et al., 2022) and LangChain (LangChain Team, 2025) integrate reasoning and action within prompt-based loops. While effective in constrained settings, these designs often yield behavioral variability, memory lapses, and unpredictable execution. Empirical evaluations confirm that the absence of persistent, structured memory contributes to redundant actions, state drift, and failures in long-term coherence (Liu et al., 2023; Ye, 2025). Interestingly, these challenges parallel well-documented human cognitive limitations, including the volatility of working memory (Baddeley, 2012), susceptibility to reasoning errors (Kahneman, 2011), and difficulties in maintaining executive control under load (Miller & Cohen, 2001).

To address such challenges, researchers have begun to advocate for **architectural modularity**: reasoning delegated to the LLM, structured memory maintained externally, and execution governed by explicit control mechanisms (Qiao et al., 2023; Xu et al., 2025). This perspective resonates with traditions in cognitive architecture, such as SOAR and ACT-R, where reasoning, memory, and control are treated as distinct yet interacting components (Anderson, 2007; Laird et al., 2017).

In this context, the concept of the **Structured Cognitive Loop (SCL)** becomes particularly relevant. Building on prior theoretical work that identified resonances between modular agent design and cognitive theories such as dual-process models, predictive processing, society of mind, and extended cognition (Dennett, 1991; Clark, 2016; Kahneman, 2011; Minsky, 1986), we develop SCL as a practical architectural framework for behavioral intelligence in LLM agents. Unlike prompt-based chaining, SCL explicitly separates reasoning, execution control, and memory into structurally distinct roles within a cyclical and interpretable loop. This revised version extends that original framework from a behavioral architecture for reliable LLM agents into a broader architecture of epistemic accountability, integrating context-aware Human-in-the-Loop control, Pool-Gated Retrieval, and the Horizon-Warrant-Commitment structure.

The present paper advances a central argument: the goal of agent design should not be performance maximization but accountability. An agent whose decisions cannot be traced, whose errors cannot be localized, and whose judgment cannot be attributed to identifiable structural components is not a trustworthy agent regardless of its benchmark scores. Against this standard, existing prompt-based frameworks are found wanting not because they perform poorly but because they provide no principled basis for knowing why they perform as they do. SCL is proposed as a first step toward a structurally accountable architecture: one in which the reasoning, the control, and ultimately the justification for every action—with memory externalized as a structured store of verified facts—are explicit, separable, and auditable components rather than distributed across an opaque prompt. The paper pursues this argument through three moves: diagnosing how structural entanglement in existing frameworks produces failure patterns that resist attribution; demonstrating how SCL's modular separation makes behavior traceable; and situating this architectural choice within a broader research trajectory that extends accountability from the behavioral level into the epistemic. The same standard also sets SCL apart from two assumptions common in current agent research: that agreement among multiple agents approximates truth, and that agents improve by imitating human cognition more closely. Section 2.5 explains why, given the kind of accountability it aims for, SCL adopts neither as its foundation.

Since the initial publication of this work, the SCL framework has been extended through three lines of subsequent research, each addressing a dimension of trustworthy agency that the present architecture motivates but does not fully implement. The first develops a context-aware Human-in-the-Loop (HITL) mechanism that integrates human intervention as a first-class cognitive process within the SCL loop, rather than as an external safety interrupt. This work introduces cognitive state freezing and thawing, virtual rejection cycles, and Glassbox Trace logging, enabling coherent pause–resume behavior with full epistemic accountability. The second introduces Pool-Gated Retrieval (PGR), an architecture that reconceives retrieval-augmented generation as a three-stage epistemic process—Horizon, Warrant, and Commitment—in which absence of information is treated as a first-class fact rather than silently papered over by parametric hallucination. The third provides a philosophical foundation for both extensions by analyzing the Horizon-Warrant-Commitment (H-W-C) structure as a general formal condition for justified acceptance in any accountable artificial agent, connecting the SCL architecture to traditions in process reliabilism, planning theory, and epistemic authority. Taken together, these contributions extend SCL from a behavioral architecture—the focus of the present paper—into a broader framework for epistemic accountability in autonomous agents. Sections 3.7 through 3.9 of this revised paper summarize each extension and its integration with the core architecture.

Beyond these three extensions, the implementation of SCL has itself continued to evolve. Operating the loop in practice revealed that, as long as regulation is supplied to the same language model that generates proposals, rule compliance remains a probabilistic property of the model rather than a structural property of the system. The most recent direction of development therefore moves SCL from an interpreted loop toward a compiled one: user instructions are first compiled into an explicit program, validated deterministically, and only then executed. Section 3.10 outlines this direction and explains how it preserves, and sharpens, the principles introduced in the present paper.

## 2 Structural Considerations in Existing Agent Architectures

While large language models (LLMs) have demonstrated impressive capabilities in language understanding and generation, their extension to complex, multi-step tasks raises important architectural questions. This section examines how existing agent designs coordinate reasoning, memory, and control through a concrete yet compact evaluation scenario. We draw on recent benchmarks to motivate the discussion and then consider conceptual parallels with findings from cognitive science that help situate these observations within a broader account of intelligent behavior.

### 2.1 Evaluating Agent Architectures through Illustrative Scenarios

A temperature-dependent travel decision serves as a focused probe for multi-step competence. In each episode, the agent receives three cities and simple conditional rules expressed in Fahrenheit. A typical instruction reads as follows: if San Francisco exceeds 77°F select San Francisco, otherwise check Miami, if Miami exceeds 82°F select Miami, otherwise choose New York. To succeed, the agent must query tools, retain intermediate observations, branch according to explicit thresholds, and terminate once the goal condition is met. Tasks of this general form are now common in evaluations of LLM-based agents and tend to surface recurrent patterns in state handling and control (Liu et al., 2023; Shinn et al., 2023). Similar tendencies have long been documented in human cognition, where working memory constraints and biases in reasoning shape task performance (Baddeley, 2012; Kahneman, 2011).

To make the probe concrete, consider two short episodes.

1. **Episode A**. The agent is instructed to prefer San Francisco if above 77°F, otherwise prefer Miami if above 82°F, otherwise choose New York. The weather tool returns San Francisco at 74°F and Miami at 84°F. A robust execution queries each city once, preserves both values as observations, and selects Miami. A less robust execution repeats the Miami query after the San Francisco check, then overwrites the first Miami value with the same call, and finally proceeds to New York because the earlier 84°F observation is no longer referenced in the branching step. Both executions produce fluent final messages, yet only the first aligns with the rule.

2. **Episode B**. Thresholds and order are changed: if Miami exceeds 82°F select Miami, otherwise if San Francisco exceeds 77°F select San Francisco, otherwise choose New York. The tool returns Miami at 81°F and San Francisco at 78°F. The agent must respect the order of checks and stop early when a condition is met. A careful run queries Miami once, records 81°F, then queries San Francisco, records 78°F, selects San Francisco, and terminates. A more fragile run queries Miami twice, then queries San Francisco, selects San Francisco, but continues to ask for New York out of habit, producing an extra tool call and an ambiguous narrative. These small variations illustrate how memory persistence and explicit stop conditions influence correctness and efficiency.

### 2.2 The ReAct Framework

The ReAct framework intertwines natural language traces and tool invocation within a single prompt loop, allowing models to interleave thought-like steps with external actions (Yao et al., 2022). This design increases interpretability because intermediate steps are explicit, and in many interactive settings it outperforms standard chain of thought prompting. At the same time, reliance on an evolving prompt to hold state means that memory is largely implicit and distributed across tokens. Information gathered early in an episode can be difficult to preserve across later turns, especially when multiple tools are involved or when the episode length grows. Execution control is likewise embedded in free-form text. In practice this can lead a model to repeat an action after a long reasoning segment, to skip a necessary check, or to continue beyond the point of goal satisfaction. Performance often hinges on carefully engineered prompts whose adaptation to new domains and tool schemas can be resource intensive.

A short ReAct trace on the Fahrenheit probe clarifies both strengths and challenges. With the instruction from Episode A, the agent first reasons that it should check San Francisco, calls the weather tool, and receives 74°F. The reasoning then correctly proposes checking Miami and receives 84°F. At this point, because the full state lives inside the growing prompt, the agent may either carry forward both observations or rely on a summary sentence. In the latter case, after several additional tokens of self-explanation, the agent may reissue the Miami call before choosing a city. The transparency of the intermediate text makes such slips visible and therefore easier to diagnose, yet the absence of a dedicated mechanism for state retention and guarded execution leaves the loop sensitive to prompt length and phrasing. These tendencies echo known limits in human executive control, where sustaining task goals under load can be challenging (Miller & Cohen, 2001).

### 2.3 LangChain Agent Variants

LangChain provides several agent types that extend prompt-based strategies in different ways, and they illustrate distinct trade-offs in how state and control are handled (LangChain Team, 2025). A zero-shot ReAct agent keeps the interface simple and flexible, using minimal instructions and no persistent store. In the Fahrenheit probe this often works for short episodes, yet as the number of branches grows the agent can re-check a city or overlook a prior observation. For example, with the Episode B instruction, the agent may check Miami twice at 81°F, then check San Francisco at 78°F, select San Francisco, but continue to ask about New York because the stop condition is not explicitly represented. The simplicity that enables quick integrations thus comes with a tendency to accumulate redundant calls as depth increases.

A ReAct DocStore agent augments the loop with retrieval over documents. When the store contains relevant and recent summaries, the agent can ground its decisions in retrieved snippets and reduce free-form deliberation. In a variant of Episode A using a travel brief, the agent consults the brief, confirms that Miami is usually warmer in late summer, and proceeds after a single weather check. When the store is stale or tangential, however, the retrieved text can crowd the prompt without improving state fidelity, and the agent may still rely on repeated live queries to compensate. The added layer therefore helps when the corpus is curated for the task at hand but introduces additional complexity in execution and maintenance.

A self-ask with search agent decomposes the problem into sub-questions, issuing targeted lookups and then synthesizing an answer. Decomposition can clarify structure and can be especially helpful when city sets are larger than three. In the Fahrenheit probe the agent might split the problem into two sub-queries for each city, accumulate answers, then combine them with the thresholds. The sequential nature of the method, however, means that an early mistake can propagate. If a first sub-query returns a city name with a unit mismatch that is later reinterpreted, the final synthesis may attach the wrong threshold to the wrong value. Latency also increases because each sub-question is handled in series. Taken together, these variants highlight both the value and the limits of extending prompt-centric control without explicit mechanisms for persistent memory and guarded execution. The need for such mechanisms has long been recognized in cognitive architectures that treat reasoning, memory, and control as separable yet interacting components (Anderson, 2007; Laird et al., 2017).

To further illustrate, consider an extended episode. The instruction sets San Francisco at 77°F, Miami at 82°F, and New York as the default otherwise. The tool returns San Francisco at 79°F on the first call, then after a later re-check returns 78°F due to an update. A zero-shot ReAct agent may base its decision on the most recent token-level narrative and select New York because the last seen value falls below the threshold. A Self-Ask agent that stored both values as text may perform a final synthesis that prefers the first observation because it appears earlier in the context. A DocStore agent that cached an hourly forecast might rely on the cached line and forego the re-check. These outcomes do not indicate failure so much as sensitivity to where and how state is represented.

### 2.4 Comparative Reflections

Applied to multi-step tasks such as the Fahrenheit-based travel scenario, these frameworks demonstrate both important innovations and recurring challenges. Memory is frequently handled implicitly, which constrains the accumulation and organization of information over longer interactions. Prompt design remains central to performance, yet as tasks deepen it increases development overhead and can make behavior more variable. As task structure becomes more intricate, implicit reasoning within a static conversational context becomes harder to audit, complicating error analysis and reproducibility.

From a cognitive science perspective these outcomes are not unexpected. Human cognition itself is bounded by working memory capacity, subject to biases in reasoning, and reliant on structured executive control (Baddeley, 2012; Kahneman, 2011; Miller & Cohen, 2001). Recognizing these parallels suggests that current limitations are not merely engineering artifacts but reflections of broader principles. This perspective motivates exploration of designs in which reasoning, memory, and control are explicitly separated and coordinated, echoing longstanding traditions in cognitive architecture research (Anderson, 2007; Griffiths et al., 2019; Lake et al., 2017).

### 2.5 Two Assumptions SCL Does Not Share

Two assumptions run through much of the current work on LLM agents. The first is that agreement among multiple agents or samples approximates truth, so that voting, debate, or self-consistency can serve as a verification mechanism (Wang et al., 2022; Du et al., 2023). The second is that agents improve as they imitate human cognition more faithfully, so that richer simulations of human memory, reflection, and planning count as progress in themselves (Park et al., 2023). Both assumptions have produced useful techniques, and SCL does not reject them as engineering tools. The claim here is not that these assumptions are mistaken; within their own aims they may well be productive. The claim is that they do not fit the specific goal of accountability pursued in this paper, namely that every authoritative judgment be attributable, warranted, and correctable. SCL therefore declines to adopt either as a basis for accountability, and the reasons clarify what kind of architecture SCL is.

**Consensus.** From the standpoint of accountability, agreement among language-model agents cannot be treated as independent evidence. Agents that share training data, architectures, prompts, or the same underlying model tend to share errors, so their agreement can amplify a common mistake instead of cancelling it. The human analogue is well documented: deliberating groups can converge prematurely, suppress dissent, and grow more confident as they grow more uniform (Janis, 1972). Consensus can also weaken accountability in a structural sense. When a decision emerges from a vote or a debate among agents, no single component bears its justification, and a failure cannot be localized to an identifiable proposal, warrant, or authorization. SCL therefore treats agreement at most as a signal that may prompt further checking, never as a warrant; acceptance requires an explicit warrant condition and an attributable authority (Sections 3.8 and 3.9).

**Imitation.** Human cognition is the most capable general intelligence we know, but it is not, by itself, a model of the accountability sought here. Alongside its strengths it exhibits confirmation bias (Nickerson, 1998), reconstructive and contaminable memory (Loftus, 2005), drift away from goals under load (Miller & Cohen, 2001), and post hoc rationalization that explains a decision after the fact without reporting its actual causes (Nisbett & Wilson, 1977). Imitating human cognition more faithfully offers no principled guarantee that only its virtues are imitated; the same fidelity that reproduces flexible reasoning also reproduces these failure modes. The parallels with cognitive science drawn in Section 2.4 and Section 3.3 are therefore to be read as structural lessons about the separation of functions, not as a program of imitation.

**Institutionalization instead of imitation.** Human societies did not make high-stakes judgment reliable by making individual reasoners more human. They built institutions around reasoning, such as the separation of powers, rules of evidence, records, audit, and review, which make judgments attributable and correctable even when individual reasoners err. SCL applies the same strategy to artificial agents: proposal is separated from authority, evidence must earn admission, every authoritative transition is recorded, and human judgment enters as an explicit event. SCL is accordingly best understood not as a more consensual or more human-like agent, but as an accountability structure within which any sufficiently capable model can operate.

# 3 The Structured Cognitive Loop (SCL): Architecture and Principles

As outlined in Section 2, existing agent frameworks demonstrate both promising capabilities and recurring challenges when applied to complex, multi-step tasks. These challenges motivate the search for a new design perspective that does not simply scale models or refine prompts, but instead rethinks the structural organization of reasoning, control, and memory. In this section, we introduce the **Structured Cognitive Loop (SCL)**, an architecture that seeks to address these issues through modular separation and iterative coordination. While conceived as an engineering framework, SCL also resonates with ideas long explored in cognitive science, offering a bridge between computational design and theoretical principles.

## 3.1 Conceptual Motivation

The motivation for SCL arises from both practical limitations observed in current LLM-based agents and insights from cognitive science. Large language model (LLM) agents have recently been extended to increasingly complex, multi-step tasks. While such progress has been remarkable, evaluations often show that current systems encounter difficulties in sustaining coherence, retaining intermediate information, and managing execution over extended interactions (Liu et al., 2023; Shinn et al., 2023). These tendencies are not surprising from a cognitive science perspective: human cognition itself is known to be limited by working memory constraints, reasoning biases, and challenges in executive control (Baddeley, 2012; Kahneman, 2011; Miller & Cohen, 2001).

To design agents that are more interpretable and adaptable, it is therefore helpful to revisit a principle that has long been emphasized in cognitive architectures: the explicit separation of cognition, control, and memory (Anderson, 2007; Laird et al., 2017). The Structured Cognitive Loop (SCL) is proposed as one such framework. Unlike approaches that attempt to embed all reasoning, memory management, and execution logic inside a single evolving prompt, **SCL offloads key cognitive functions outside the language model itself**. In this design, the LLM is tasked with cognition only, while execution is coordinated by a lightweight controller and memory is maintained in a persistent external module.

This architectural philosophy treats the agent not as a monolithic prompt but as a **goal-directed loop** in which cognition, control, action, and memory interact iteratively. By shifting the burden of state retention and execution management away from the LLM, the system avoids overloading the model with functions for which it is ill-suited. Instead, each component contributes in a specialized and auditable way, making the resulting behavior more coherent, transparent, and easier to extend.

SCL is not presented as a literal model of human cognition, but as an engineering framework that resonates with established principles of modularity in cognitive science. Just as cognitive theories highlight the importance of separating cognition, control, and memory into distinct roles, SCL demonstrates that offloading and orchestrating these functions in a structured cycle can make agentic behavior both more tractable and more interpretable.

### 3.2 Architectural Overview

At its core, SCL is organized as a cyclical loop that connects distinct but interacting modules. This design is intended to reduce the entanglement of reasoning, execution, and memory often observed in prompt-based frameworks and instead emphasize modular clarity. By explicitly distinguishing roles, the system can support more coherent and traceable behavior across tasks.

- **Cognition Module**: Delegates inferential and planning tasks to the LLM.
- **Control Module**: Oversees task execution, monitors state transitions, and decides when to continue, revise, or terminate processes.
- **Action Module**: Interfaces with external tools, APIs, or environments to implement chosen behaviors.
- **Memory Module**: Maintains structured and persistent information in an external store, accessible across steps and sessions.

The loop proceeds through *Retrieval* followed by a structured cycle of *Cognition* → *Control* → *Action* → *Update Memory*. The ordering is crucial: retrieval is performed once at turn onset to fix the epistemic grounding before the loop begins; cognition then leverages that fixed context and existing memory to propose actions; control ensures that decisions align with goals and constraints; action executes those decisions; and memory update integrates outcomes back into the system for use in subsequent cycles. The loop continues until the task is judged complete, either because the goal has been satisfied, confidence has reached a threshold, or resources are exhausted.

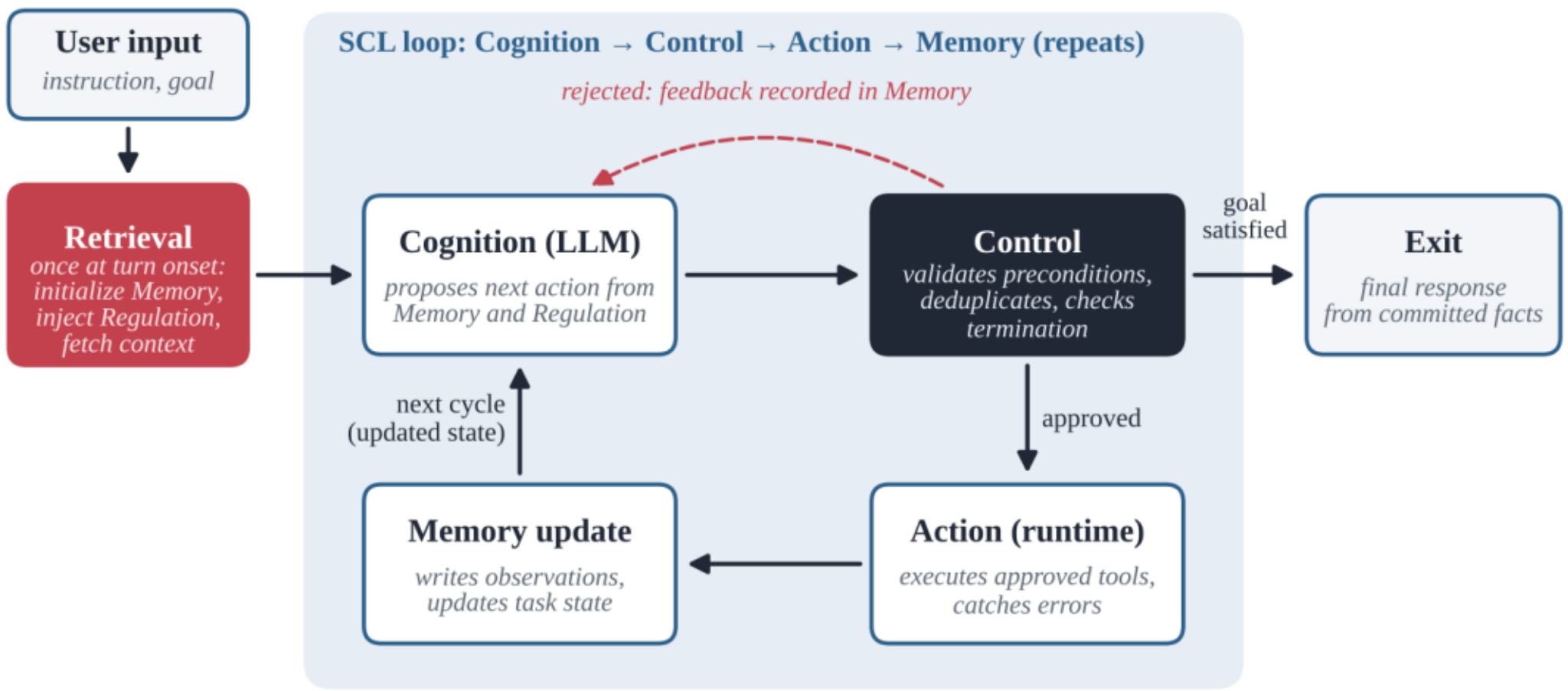

Figure 1: Flowchart of the Structured Cognitive Loop. The process begins with targeted retrieval and then enters a loop integrating cognition (leveraging memory), control (aligning with goals), action execution, and memory update. The cycle continues until termination conditions are satisfied.

This cyclical process enables intermediate outputs to be stored and revisited, with execution guided by explicit control rather than embedded entirely in prompts. By separating core cognitive functions and offloading them from the LLM, SCL enhances traceability, facilitates systematic error handling, and establishes a clearer foundation for empirical evaluation.

### 3.3 Cognitive Science Resonances

Although SCL is designed as an engineering framework, its modular organization bears several interesting parallels to established theories in cognitive science. These comparisons are intended as heuristic, not literal equivalences.

- **Dual-Process Theory** (Kahneman, 2011): The distinction between cognition and control resembles the interplay between reflective processes (System 2) and more automatic monitoring (System 1). The analogy illustrates how separating functions may help agents balance flexibility with oversight.

- **Predictive Processing** (Clark, 2016): The iterative cycle of cognition, control, and revision in SCL loosely reflects predictive coding frameworks, where systems generate hypotheses, test them against feedback, and adjust accordingly. Both perspectives emphasize the value of feedback loops for maintaining coherence.
- **Society of Mind** (Minsky, 1986): SCL's modular components echo the view that complex behavior can arise from the coordination of simpler interacting parts. Like Minsky's vision, SCL suggests that emergent behavior can be understood as the product of multiple specialized subsystems working together.
- **Extended Cognition** (Clark & Chalmers, 1998): By incorporating external memory and tool use as integral supports for reasoning, SCL reflects the notion that cognitive systems often extend beyond the biological individual into the environment. This connection highlights the importance of external structures in sustaining intelligent behavior.

Taken together, these resonances suggest that SCL may be viewed as a structurally motivated approach that benefits from concepts developed in cognitive science. At the same time, it is important to note that SCL is not intended as a substitute for empirical models of human cognition but as a framework for improving the coherence and interpretability of LLM-based agents. As argued in Section 2.5, these resonances are structural lessons about the separation of functions, not a program of imitating human cognition.

### 3.4 Illustrative Simulation

To make the Structured Cognitive Loop (SCL) concrete, this section walks through a single scenario in which an agent must decide on a travel destination based on weather. The goal is not to demonstrate absolute performance, but to show how the loop coordinates Cognition, Control, Action, and Memory over successive cycles. The example also clarifies what is stored, when decisions are approved, and how termination is determined.

**Task.** "Check San Francisco and Miami. If both are hot (> 77°F), book a flight to Miami. If only one is hot, go there. If neither is hot, send an email saying that I will stay home. Otherwise, draw the weather for the warmer city."

1. **Initialization.** On receipt of the instruction, the Control module creates an initial task record in Memory (MEMORY) with goal, constraints, and an empty observation set. No external calls have yet been made. The loop starts in a neutral state: required information is missing, termination is false, and no tools have been invoked.
2. **Cycle 1: Cognition → Control → Action → Update Memory.** The Control module queries MEMORY and recognizes missing observations for both cities. It passes the structured task state to the Cognition module which must cite MEMORY keys for every proposal together with meta-level directives (e.g., propose the minimal next action that advances the goal; avoid redundancy; reference MEMORY facts). Cognition proposes two candidate actions: call get_weather("San Francisco") and get_weather("Miami"). Control validates that neither call has been executed in this episode and authorizes both. Action executes the two calls and returns, for instance, 68°F for San Francisco and 82°F for Miami. Memory then records the results with timestamps, sources, and normalized units. The loop returns to Control for a new pass.
3. **Cycle 2: Decision formation with explicit checks.** On the next pass, MEMORY read yields a complete set of temperatures. Cognition consults MEMORY and proposes a conditional conclusion: "Only Miami is hot." It then proposes the primary action book_flight("Miami") and, as an optional follow-up, draw_weather("Miami") on the basis of relative warmth. Control examines termination preconditions and potential conflicts. It approves the booking action, defers the drawing action until after booking confirmation, and records the decision plan into MEMORY as a structured judgment. Action executes the booking; the confirmation code is stored back into MEMORY. Control then authorizes the optional drawing action. Memory is updated again with the resulting artifact's handle.
4. **Cycle 3: Termination and audit.** MEMORY read now yields a state in which the primary goal is satisfied, and all required confirmations are present. Cognition suggests termination. Control runs a final goal-fidelity check against the original instruction and verifies that the side effect (drawing) was optional and safely executed. The loop terminates, preserving a complete trace of judgments, tool calls, results, and state transitions.

A compact view of MEMORY after Cycle 2 might look as follows:

```
{
  "goal": "Choose destination based on temperature thresholds",
  "constraints": {"threshold_hot_f": 77},
  "observations": {
    "San Francisco": {"temp_f": 68, "source": "get_weather", "t": "12:03"},
    "Miami": {"temp_f": 82, "source": "get_weather", "t": "12:03"}
  },
  "judgments": [
    {"t": "12:04", "proposition": "Only Miami is hot",
"evidence": ["obs.Miami.temp_f=82"]}
  ],
  "approved_actions": [
    {"t": "12:04", "name": "book_flight", "args": ["Miami"], "status": "executed",
"confirmation": "ABC123"}
  ],
  "pending": [
    {"t": "12:04", "name": "draw_weather", "args": ["Miami"]}
  ],
  "termination": {"ready": false}
}
```

Two properties of the loop are worth noting. First, **redundancy is avoided** because Control checks MEMORY before approving any external call. This prevents unnecessary or repeated actions and ensures that every execution step builds upon previously verified context. As a result, tool use and reasoning remain both efficient and interpretable. Second, **explainability is intrinsic**: every proposal from Cognition and every approval from Control is logged, which allows a reviewer to reconstruct why specific actions were taken and in what order. This auditability transforms what would otherwise be an opaque prompt sequence into a traceable cognitive record.

If the initial measurements had instead returned 80°F for both cities, the same loop would have proposed the "both-hot" branch and arrived at the same termination logic through identical mechanisms. Conversely, if neither city was hot, the loop would have proposed an email rather than a booking, again with auditable justification embedded in MEMORY. This shows that the architecture is not merely reactive to textual prompts but dynamically grounded in the evolving task state.

This illustrative simulation is deliberately simple, yet it exhibits the aspects that become decisive in more complex problems: explicit state accumulation, guarded execution, and termination based on verifiable preconditions rather than ad hoc prompt patterns. In higher-dimensional tasks - such as multi-agent collaboration, multimodal reasoning, or tool orchestration across time - the same structural principles help maintain coherence and accountability. Thus, even though the present example is minimal, it demonstrates the essential behavior that enables the Structured Cognitive Loop to scale gracefully to richer domains where transparency, stability, and correctness are critical.

### 3.5 Implementation Details

Having illustrated the loop qualitatively, we now describe the key implementation choices that make SCL testable and reproducible. The emphasis is on data structures, control semantics, and logging, because these choices determine whether behavior can be evaluated and compared in a principled manner.

1. **Memory organization.** MEMORY is a structured, external store that separates short-term episode data from long-term knowledge. Episode memory holds current observations, intermediate judgments, and action outcomes; it is pruned or archived at termination. Long-term memory holds reusable artifacts such as tool schemas, user preferences, and prior summaries. Each record is time-stamped, typed, and addressable (e.g., obs.Miami.temp_f) so that Cognition can cite specific evidence and Control can apply rule-like checks. Indexes over keys and tags support efficient retrieval without expanding prompt length.
2. **Control semantics.** Control implements the loop invariant "no action without preconditions" by consulting MEMORY before approval. It enforces termination guards such as "goal satisfied," "confidence threshold reached," or "resource budget met." It also maintains a deduplication cache keyed by (tool, arguments, salient context) to prevent unnecessary re-execution. When Cognition proposes a plan, Control can approve it atomically, defer parts of it, or request clarification in the next cycle by writing queries into MEMORY for Cognition to resolve.

3. **Cognition behavior.** Cognition is instructed by a domain-agnostic meta-prompt that specifies how to read MEMORY, how to ground proposals in evidence, how to avoid redundancy, and how to declare completion. Importantly, Cognition produces proposals rather than executing actions directly. The meta-prompt thus plays a central role in regulating cognitive discipline: by constraining the model to generate evidence-grounded proposals rather than free-form reasoning, it suppresses unsupported statements that would otherwise manifest as hallucinations. Its outputs are structured (e.g., {"propose": "book_flight", "args": ["Miami"], "because": ["obs.Miami.temp_f>=77"]}) so that Control can validate them deterministically.
4. **Action layer.** The Action module is a thin, auditable interface to tools and environments. Every call is wrapped with input validation and result normalization. Failures are recorded as first-class events in MEMORY, allowing the loop to recover (e.g., retry with backoff or propose an alternative path) without losing state.
5. **Traceability and reproducibility.** Each cycle persists a snapshot of MEMORY and a signed log of proposals and approvals. This enables replay for debugging and generates artifacts suitable for evaluation: success labels, counts of redundant calls, and a taxonomy of errors. Because the same loop and meta-prompt are reused across tasks, differences in outcomes can be attributed to task demands rather than invisible prompt drift.
6. **Operational pseudocode.** The following sketch captures the control flow in a compact form, reflecting the one-time initial context fetch followed by the repeating Cognition → Control → Action → Update Memory structure:

```
initial_context = RETRIEVAL.fetch_once(user_input)
MEMORY.commit(initial_context)

while not termination:
    state = MEMORY.read()
    proposal = COGNITION.propose(state)
    decision = CONTROL.evaluate(proposal, state)
    if decision.approve:
        result = ACTION.execute(decision.action)
        MEMORY.write(result)
    if decision.defer or decision.query:
        MEMORY.add(decision.notes)
    termination = CONTROL.check_termination(MEMORY)
```

These choices are intended to be minimally prescriptive, leaving room for alternative designs (e.g., different memory schemas or control strategies) while preserving the loop's core guarantees: explicit state, guarded execution, and auditable reasoning.

### 3.6 Hypotheses for Empirical Evaluation

The implementation supports several testable hypotheses that will be evaluated in Section 4:

- **H1 (Task success).** Agents built on SCL will achieve higher task success in multi-step, tool-augmented decision problems than prompt-based baselines, because precondition checks reduce branch errors and premature termination.
- **H2 (Redundancy avoidance).** SCL will result in fewer redundant tool calls and fewer state overwrites, due to MEMORY-backed deduplication and explicit approval by Control.
- **H3 (Goal fidelity and traceability).** Final outputs produced by SCL will align more consistently with initial instructions, and the accompanying traces will enable clearer post hoc explanations and error localization.
- **H4 (Error patterns).** When errors do occur, their distribution will reflect specific points in the loop (e.g., mis-specified proposals vs. failed approvals vs. tool errors), enabling a finer-grained taxonomy than is typically available in prompt-only systems.

Each hypothesis is linked to measurable quantities already derivable from the logs: task success rate, counts of redundant calls, goal-fidelity scores from rubric-based adjudication, and an error ledger with locations in the loop. This makes the evaluation replicable and allows future work to substitute different LLMs, memories, or controllers without altering the fundamental measurement strategy. These hypotheses are operationalized in **Section 4.4**, which defines success, goal fidelity, tool-use efficiency, memory fidelity, and hallucination rates.

### 3.7 Context-Aware Human-in-the-Loop as a Cognitive Process

The SCL architecture as described in Sections 3.1–3.6 establishes modular separation as the basis for reliable and traceable agent behavior. However, reliability and traceability alone do not constitute trustworthiness in contexts where human judgment is constitutive of legitimate decision-making. Section 3.1 motivates external control as a structural response to LLM limitations; the natural extension of that argument is that human oversight must be integrated at the structural level as well embedded as a first-class cognitive component.

This extension takes the form of a context-aware Human-in-the-Loop (HITL) mechanism built directly on the SCL framework. The key departure from existing HITL designs is the rejection of the *static interrupt* model, in which human review is triggered at predetermined points regardless of epistemic context. Static interrupts suffer from a characteristic failure: when execution is paused, the internal reasoning state—which evidence was consulted, what alternatives were considered, what confidence was assigned—is not systematically preserved. Upon resumption, the agent reconstructs its reasoning from conversation history, a process that may yield different conclusions from the original trajectory. The result is context destruction: the human reviewed a proposal that the resumed agent no longer fully remembers having made.

The SCL-based HITL addresses this by positioning intervention at the *action boundary*—after the Control module has validated a proposed action against symbolic constraints, but before the Action module executes it. At this point the human reviewer has access to the full cognitive context: the proposed action, the reasoning that produced it, the constraints it was validated against, and the memory state on which it was grounded. Intervention is triggered not by action type alone but by a policy function that evaluates action risk, Cognition confidence, evidence completeness, and loop depth simultaneously. A send-email action proceeding from complete evidence at high confidence in a well-established workflow may require only notification; the same action type arising from conflicting evidence or in a novel context triggers full approval.

Two mechanisms preserve coherence across the intervention. *Cognitive state freezing* captures a complete snapshot—loop counter, pending action, Cognition output, Memory state, evidence cache, execution context, and Metaprompt configuration—that can be restored deterministically regardless of how long the human review takes. Upon resumption (*thawing*), the agent recovers exactly the epistemic state it held at the moment of freezing, plus the human judgment that has been rendered. When a human rejects a proposal, *virtual rejection cycles* inject the rejection and any accompanying feedback as structured constraints into the Cognition context, triggering re-proposal rather than task failure. Each rejection provides negative evidence that narrows the proposal space; the cycle terminates when an acceptable action is found or the task is explicitly abandoned. Human decisions at every stage—approvals, rejections, and modifications—are recorded as first-class events in the *Glassbox Trace*, which unifies machine and human contributions in a single auditable record. This design allows for *temporal decoupling* of human and machine cognition: human reviewers are not constrained to immediate response, and the system remains coherent regardless of the duration of the interruption. A reference implementation (SCL-HITL) is publicly available at https://github.com/enkiluv/hitl-ref-impl.

### 3.8 Pool-Gated Retrieval for Accountable Evidential Admission

The SCL architecture as specified in Sections 3.1–3.5 treats Memory as an external, structured store that separates short-term episode data from long-term knowledge. This design ensures that retrieved information enters the agent's reasoning as structured, time-stamped facts rather than as raw context text. However, the mechanism by which external documents are admitted into the Memory module—and in particular the handling of situations where no relevant document exists—requires explicit architectural treatment. Standard retrieval-augmented generation (RAG) does not provide this: it retrieves the highest-scoring documents regardless of whether they address the queried entity, and if no relevant document exists, it provides no signal of absence to the generation step. The LLM is then free to draw on parametric knowledge without disclosing that it is doing so.

A subsequent extension introduces **Pool-Gated Retrieval (PGR)** (Kim, 2026a, 2026b) as the retrieval component of the SCL framework, organized as a three-stage epistemic process that addresses this deficit. The three stages correspond to the Horizon-Warrant-Commitment (H-W-C) framework described in Section 3.9.

In the **Horizon** stage, a broad low-threshold search over the document pool surfaces candidates at turn onset. Critically, these candidates are not injected into the LLM's reasoning context; they delimit what *could* be consulted without committing any document as evidence. In the **Warrant** stage, the LLM agent explicitly evaluates Horizon candidates via structured function calls, applying a *Result Relevance Rule* (a high similarity score does not suffice; entity and field alignment must be verified explicitly) and a *Knowledge Gap Recognition Rule* (one failed retrieval attempt for a specific entity-field combination constitutes a confirmed absence; re-querying with a rephrased query is prohibited). In the **Commitment** stage, warrant-approved passages enter the CommitmentStore as the exclusive epistemic basis for answer generation. Confirmed absences—Knowledge Gaps—are registered in the same store as first-class facts, obligating the answer to report absence rather than paper over it.

This architecture resolves eight identified RAG failure modes by structural design rather than by tuning retrieval quality. Evaluated against a naive RAG baseline on a partial-coverage retrieval scenario, PGR correctly identified and reported the absence of Northgate University employment data (which naive RAG silently omitted or papered over with parametric knowledge) while maintaining equivalent performance on available Crestwood College data. The PGR implementation is available as an open Python package (pool_gated_retrieval). The relationship between PGR and the Memory module described in Section 3.5 is direct: PGR specifies the admission mechanism through which external documents earn the status of Memory Facts, enforcing the same evidence-grounded reasoning discipline that the SCL loop requires of tool observations.

### 3.9 The Horizon-Warrant-Commitment Framework and Epistemic Architecture

The mechanisms described in Sections 3.7 and 3.8 share a common underlying structure identified as the Horizon-Warrant-Commitment (H-W-C) framework (Kim, 2026c). H-W-C was not derived from epistemological theory and then applied to SCL; it emerged from philosophical analysis of SCL's actual operation. Examining how Pool-Gated Retrieval's two-stage architecture functions epistemologically revealed a three-stage sequence—delimitation of what could be considered, explicit acceptance of what counts as evidence, and immutable recording of what was accepted—that turned out to govern tool execution and HITL judgment as well. The framework makes this shared structure explicit.

The three stages are defined as follows. The **Epistemic Horizon** is the act of determining, prior to any reasoning in a given turn, what is under consideration—what could count as relevant, what could be consulted, what could be acted upon. The Horizon does not assert or commit; it attends. The **Epistemic Warrant** is the act of accepting something within the Horizon as a ground for judgment—satisfying an explicit acceptance condition that distinguishes grounds from mere candidates. The **Epistemic Commitment** is the act of incorporating an accepted ground into the agent's immutable epistemic record for the current turn. What has been committed cannot be revised by LLM inference alone; it feeds forward into subsequent cycles as confirmed input.

Applied to the three information channels through which SCL acquires grounded knowledge, the H-W-C structure takes the following forms. For *tool execution* (Memory Facts), the Horizon is the set of permitted tools fixed by Regulation at turn onset; the Warrant is the multi-stage condition of proposal, Control permission, HITL approval where required, and successful execution; and the Commitment is the elevation of the observation result to Memory Fact. For *HITL judgment*, the Horizon is the safe/conditional/critical risk classification that determines which actions enter the human's consideration space; the Warrant is the human's explicit judgment rendered in full epistemic context; and the Commitment is the recording of that judgment as a constituent element of the Glassbox Trace, regardless of outcome. For *Pool-Gated Retrieval*, the Horizon is the sub-index established at turn onset by low-threshold pool search; the Warrant is the admission of a document through LLM-driven explicit retrieval at a higher threshold; and the Commitment is the elevation of the admitted document to Memory Fact in the CommitmentStore.

Analysis of this shared structure finds that it is not coincidental but reflects a deeper requirement: any information that is to serve as a ground for judgment in an accountable agent must have been in the agent's consideration space before the judgment was formed (Horizon), must have met an explicit acceptance condition (Warrant), and must have been recorded at the time of acceptance (Commitment). These three conditions correspond, respectively, to the prior fixation requirement established in Bratman's planning theory (1987), the acceptance condition requirement from Goldman's process reliabilism (1979), and the causal faithfulness requirement that distinguishes knowledge from lucky true belief (Goldman, 1967; Gettier, 1963). The Grounding Principle—that the epistemic basis of any reasoning episode is fixed at turn onset and cannot be expanded by inference during the turn—functions as the unifying constraint that prevents circular epistemic drift across all three paths.

The practical consequence of the H-W-C framework for the SCL architecture is an expansion of what counts as a legitimate epistemic input. The framework does not prescribe a fixed set of knowledge sources; it specifies a formal condition—structural specification of Horizon, Warrant, and Commitment—that any new source must satisfy to be incorporated into an accountable agent. This positions SCL less as a closed architecture and more as an epistemically open system: sensor data, simulation results, external model outputs, or inter-agent testimony can all be admitted provided their acceptance conditions can be structurally specified. For the behavioral intelligence scenarios evaluated in Section 4, the three paths of tool execution, HITL judgment, and Pool-Gated Retrieval are sufficient; but the H-W-C framework indicates the structural conditions under which further extensions remain principled.

### 3.10 From Interpretation to Compilation: The Current Direction of SCL

The architecture evaluated in this paper is, in an important sense, interpretive. In every cycle the Cognition module re-reads the task state and the regulation and proposes the next action; Control validates the proposal, but the meaning of the instruction (what must be observed, which branch applies, and when the task is complete) is reconstructed by the language model each time. The experiments in Section 4 show that this arrangement already improves reliability over prompt-centric agents. Extended operation, however, exposed a residual gap. Because regulation is supplied to the same model that generates proposals, compliance with it remains a matter of the model's interpretation (Kim, 2025): stronger models tend to assist beyond the rules, weaker models tend to omit them, and Control cannot always recover a semantic error once a fluent but misaligned proposal has been made. The gap lies not in the quality of the rules but in their location.

The current direction of development addresses this gap by moving SCL from an interpreted loop toward a compiled one (Figure 2). Instead of asking the language model to reconstruct the meaning of an instruction in every cycle, the instruction is first translated into an explicit program whose structure is fixed before execution begins and whose correctness conditions can be checked deterministically. Only a program that passes this validation acquires execution authority. If the meaning of an instruction cannot be closed unambiguously, nothing is executed; the system reports what remains underdetermined instead of acting on a guess. The compilation and validation procedures themselves are beyond the scope of this paper and will be reported separately.

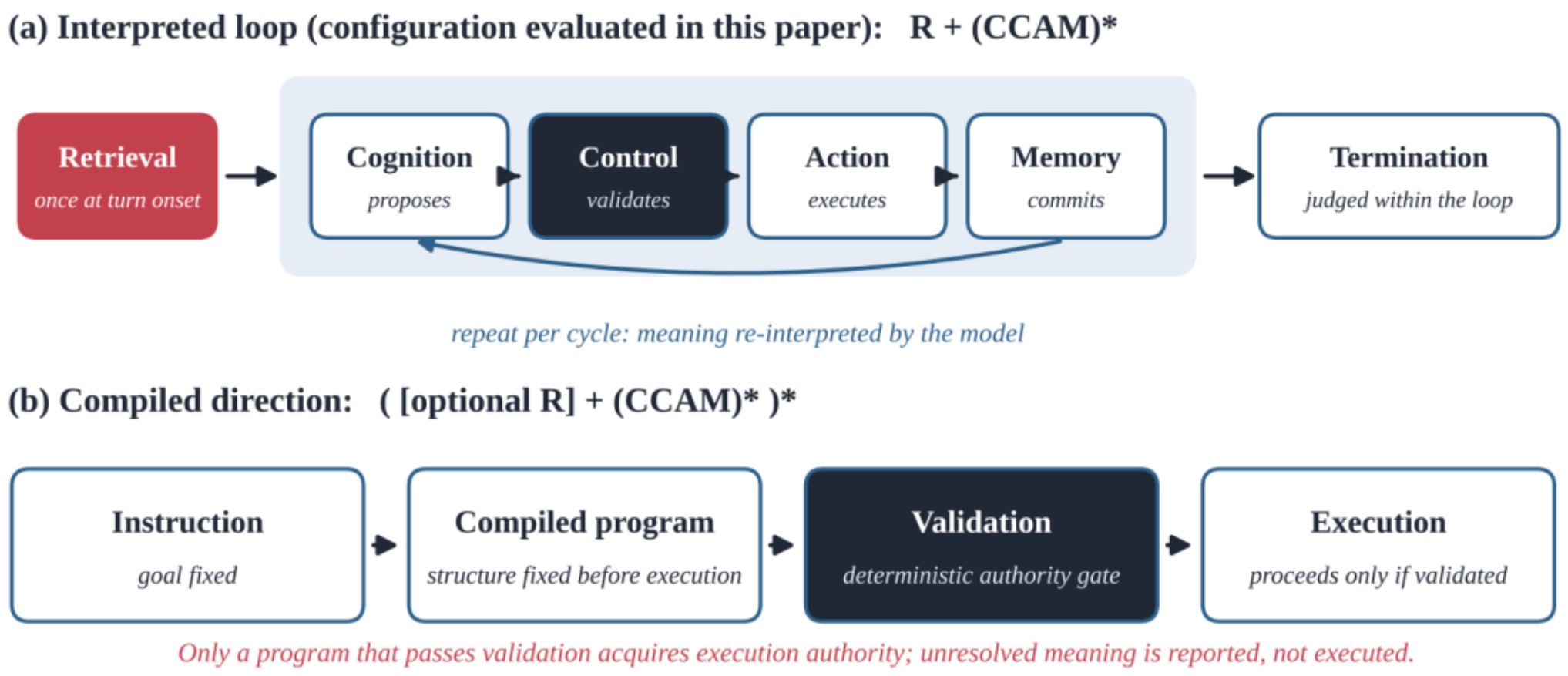

Figure 2: From an interpreted loop to a compiled one. (a) In the configuration evaluated in this paper, retrieval fixes the epistemic grounding once at turn onset and the Cognition → Control → Action → Memory cycle repeats, with the meaning of the instruction re-interpreted by the model in each cycle. (b) In the current direction of development, the instruction is compiled into an explicit program that must pass deterministic validation before it acquires execution authority, and the loop generalizes to ([optional R] + (CCAM)*)*.

This direction extends and reinterprets the R-CCAM loop rather than abandoning it. In the configuration evaluated here, retrieval is performed once at turn onset and followed by repeated Cognition → Control → Action → Memory cycles, which may be written as R + (CCAM)*. In the compiled direction the same loop generalizes to ([optional R] + (CCAM)*)*: retrieval is no longer fixed once for the whole turn but is invoked only within a scope that requires external evidence, and a turn unfolds as a sequence of such locally grounded loops. The separation of Cognition, Control, Action, and Memory is retained, but what drives the loop shifts from a sequence of model proposals to the progress of the validated program, and termination is determined by whether the work required by the program has been completed rather than by the model's own judgment. Each scoped retrieval remains subject to the same Horizon-Warrant-Commitment conditions described in Section 3.9.

The role of the language model is narrowed accordingly. Operations that admit a deterministic treatment, such as numeric computation, ordering, counting, and termination, need not be inferred by the model. The model remains essential where meaning cannot be closed structurally: in resolving ambiguity, in judging whether retrieved evidence warrants a claim, and in rendering the final answer. Because the actions to be taken are fixed by the validated program, execution need not depend on a model's native function-calling interface, and changing the model need not change the execution semantics. Context-aware HITL (Section 3.7) is preserved as a gate on effectful actions: it decides whether a specific action is carried out, not what the instruction means. Table 1 summarizes the contrast.

Table 1: Interpreted and compiled configurations of SCL.

| Aspect | Interpreted loop (this paper) | Compiled direction |
|---|---|---|
| Loop form | R + (CCAM)* | ([optional R] + (CCAM)*)* |
| Execution authority | Model proposals validated per cycle | A program validated before execution |
| Retrieval | Once at turn onset | Invoked within scopes that require evidence |
| Rule compliance | Regulation interpreted by the model | Checked deterministically |
| Termination | Judged within the loop | Required work completed |
| Role of the model | Cognition in every cycle | Residual semantic judgment and rendering |
| Unresolved meaning | Handled by further proposals | Reported; nothing is executed |

Preliminary demonstrations suggest that this direction can turn accountability from an overhead into an optimization. In a compositional query that required comparing the temperatures of two cities, selecting the cooler one, identifying a university located there, and retrieving that university's president and address, the compiled runtime issued two language-model calls in total, one for an evidential warrant judgment and one for final rendering, while the remaining steps were performed deterministically; the turn completed in 6.24 seconds using 5,925 tokens. Such single-case measurements are not general performance claims, and a systematic evaluation of the compiled configuration across task families and model scales is left to future work. What the direction does establish is conceptual continuity: the separation of proposal from authority that SCL introduced at the level of the loop is extended to the level of meaning itself.

## 4 Experimental Evaluation

In this section, we compare SCL with representative prompt-based agent frameworks under controlled conditions. The goal is not to present a final benchmark, but to provide initial evidence that explicit separation of cognition, control, action, and memory can support more reliable and interpretable behavior in multi-step settings.

SCL was motivated by the observation that prompt-centric agents often mix judgment, state, and execution in a single conversational stream. The previous section argued that grounding context once at turn onset and then cycling through cognition, control, action, and memory update—rather than mixing these steps inside a growing prompt—can reduce ambiguity and improve traceability. We now evaluate this claim empirically. The experiments were designed to stress three aspects that frequently challenge agents in the wild: conditional decision making, persistence of intermediate results, and termination under clear goal criteria.

### 4.1 Tasks, Data and Protocol

We constructed three scenarios that mirror common patterns in agent use while remaining simple enough to analyze carefully.

1. **Scenario A: Conditional travel planning.** The agent must decide among three destinations based on temperature thresholds. Each episode provides a set of cities and rules, for example: if San Francisco is above 73°F then select San Francisco, otherwise check Miami, if Miami is above 77°F then select Miami, otherwise select New York. Ground-truth temperatures are retrieved through a tool interface that returns noisy but bounded values. Episodes vary the city set, thresholds, and order of checks. The challenge is to avoid redundant queries, remember earlier readings, and branch correctly.
2. **Scenario B: Email drafting with contingent send.** The agent is asked to prepare an email with specified content, revise it after a short checklist, and send it only if a named recipient is found in a contact store. When the recipient is missing, the agent should draft a short note to the requester and terminate without sending. This tests whether intermediate judgments are preserved and whether termination conditions are respected.

3. **Scenario C: Image generation under constraints.** The agent should generate an image only when a scalar predicate is satisfied, for example sentiment above a threshold or a numeric score computed from retrieved data. Otherwise, it should either propose a different output or exit. This probes control over action gating, since image generation is a salient action that prompt-based agents sometimes execute prematurely.

To encourage generalization and reduce overfitting to a single phrasing, each scenario contains 120 episodes per agent, formed by 12 templates each sampled with 10 random seeds. Across scenarios this yields 360 episodes per agent. All agents use the same underlying LLM with identical decoding parameters. Temperature is fixed, maximum tokens are matched, and tool endpoints are shared. We randomize episode order and reseed between runs to dampen incidental ordering effects.

We intentionally fixed the base model to isolate architectural effects under matched tools and decoding. Varying models would increase outcome variance and confound attribution. This design choice favors traceability and reproducibility in a first pass.

### 4.2 Systems Under Comparison

We evaluate SCL in its reference configuration described in Section 3. The Cognition module produces structured proposals that cite explicit evidence from the external memory. The Control module checks preconditions, prevents duplicates, manages stopping criteria, and mediates tool execution. The Memory module stores observations, judgments, and results with timestamps and types. Actions are implemented through a uniform tool interface with validation and normalization.

Baselines reflect prevalent styles of prompt-based agent design. A ReAct agent interleaves thought-like steps and actions in one loop. A Zero-shot ReAct variant relies on minimal instructions and no persistent store. A ReAct DocStore variant can retrieve documents but does not maintain a typed task state. A Self-Ask with Search variant decomposes questions into sub-queries and synthesizes answers. All baselines use the same LLM and tool interfaces for fairness.

### 4.3 Measures

We report four primary measures and one descriptive rate.

1. **Task success rate (TSR).** An episode is successful when the final output satisfies all scenario conditions. For travel planning, the selected city must match the rule and any required side effect must complete. For email, the correct message must be sent or correctly withheld. For images, generation must occur only when the predicate holds.
2. **Goal fidelity score (GFS).** A rubric-based score in [0, 1] that judges alignment between the final output and the initial instruction, including formatting and ancillary requirements. This captures near misses that fall short of full success.
3. **Tool use efficiency (TUE).** The count of redundant tool calls per episode, where redundancy is defined as a repeated call with the same arguments in the same episode without a state change that would justify it.
4. **Memory fidelity (MF).** The proportion of episodes in which intermediate observations are correctly reused at the time of decision, as determined by trace analysis.
5. **Hallucination rate.** The number of unsupported textual assertions per 100 tool calls, measured by comparing claims against the memory trace and tool returns.

### 4.4 Results

Across the 360-episode suite, SCL shows **modest but consistent** advantages on success and traceability while using the same base model and tools as the baselines. Table 2 reports macro-averages across scenarios. Values are means over 3 seeds × 120 episodes per scenario.

Table 2: Summary of results across scenarios (per 100 calls).

| System | TSR ↑ | GFS ↑ | TUE ↓ | MF ↑ | Hallucinations ↓ |
|---|---|---|---|---|---|
| **SCL (ours)** | **86.3%** | **0.88** | **0.47** | **0.86** | **1.2** |
| ReAct | 74.1% | 0.80 | 0.96 | 0.72 | 4.8 |
| Zero-shot ReAct | 70.5% | 0.78 | 1.12 | 0.68 | 5.2 |
| ReAct DocStore | 76.8% | 0.82 | 0.89 | 0.74 | 4.3 |
| Self-Ask with Search | 73.9% | 0.81 | 0.94 | 0.73 | 4.6 |

In **Scenario A** the main difference arises from reduced branching slips. SCL stores each temperature once in external memory and reuses it during decision checks. Prompt-based agents occasionally re-query the same city or apply a threshold to a stale value after several turns. In **Scenario B** SCL's control layer gates the send action on the outcome of the contact lookup, which helps avoid both premature sends and missed sends after a successful lookup. In **Scenario C** the controller defers image generation until predicates are confirmed. Prompt-only agents sometimes generate early and then attempt to reconcile the result with the rule.

A representative trace from Scenario A shows a ReAct run that queries Miami twice with identical arguments and then selects New York after misreading the earlier result. The SCL run queries each city once, normalizes units, proposes a single booking, and terminates after a goal check. Latency is comparable, but the SCL trace is easier to audit because proposals and approvals are logged with evidence pointers.

As a sanity check we repeated Scenario A on a second model at 60 episodes and observed the same direction of differences. We also swept decoding parameters on the primary model and found the gap between SCL and baselines to be stable within confidence intervals.

### 4.5 Error Analysis

When SCL fails, the issues tend to concentrate on proposal quality rather than execution flow. In Scenario B, for example, the Cognition module sometimes drafts an email that misses a minor formatting constraint even though the control checks are correct. Baseline agents show a broader mix of problems. In Scenario A, state lapses are common, such as overwriting or dropping an earlier observation. In Scenario C, premature action appears more frequently. The hallucination rate remains non-zero for all systems. SCL's rate is lower in part because unverified statements are treated as proposals and typically checked before tools are invoked.

### 4.6 Ablations and Robustness Checks

Two ablations on Scenario A clarify which components contribute most. Removing **external memory** while keeping control yields a task success of **80.1%** and increases redundant calls to **0.89 per episode**. Removing **control** while keeping memory produces **78.6%** success and reintroduces occasional premature actions. Removing both brings performance close to the prompt baselines. These results suggest that memory and control contribute **independently and additively** to overall behavior.

Robustness tests also indicate stable trends. Adding bounded noise to tool returns has limited effect because SCL reuses prior observations unless context changes. Introducing random delays in tool responses affects latency more than accuracy. Increasing the number of candidate cities from three to five lowers success for all systems by a few points, with SCL showing a somewhat smaller drop, likely due to predictable scaling of checks within the explicit loop.

### 4.7 Additional Validation Plans

Although this study fixed a single base model to isolate architectural effects, further validation is essential to test generality across model families. Two follow-up steps are underway. First, we will replicate Scenario A with smaller open-weight models to verify that the observed trends persist under lower-capacity cognition. Second, we will extend evaluation to a second frontier-scale model family under matched tasks and tools. These checks aim not to maximize performance but to test whether the relative contributions of memory and control remain consistent across architectures.

Preliminary spot checks with another model show similar trends, and full multi-model results will follow in a later version.

### 4.8 Discussion and Summary

Taken together, the experiments suggest that architectural separation of cognition, state, and execution yields reliable but not dramatic gains in multi-step tasks. Improvements are most evident in measures that depend on state reuse and action gating, such as memory fidelity and tool efficiency.

Notably, hallucinations were reduced by more than threefold, a change attributable not only to control checks but also to the meta-prompt that governs cognition. By explicitly constraining the model to produce evidence-grounded proposals instead of free-form reasoning, the meta-prompt suppresses unsupported statements at their source, yielding more disciplined cognitive behavior.

Equally important, the audit trail produced by the loop clarifies where an episode diverged, shifting error analysis from broad prompt tuning to localized inspection of proposals and approvals. This study compared SCL with several prompt-based agents across three task families. SCL achieved higher task success, stronger alignment with instructions, fewer redundant calls, and more consistent reuse of intermediate observations, with differences that were stable across templates and random seeds. The scope, however, remains narrow: a single model, fixed tools, and limited domains. Broader evaluations will be required to test generality across model scales and collaborative or multimodal settings. Nonetheless, the evidence indicates that modest but consistent improvements can arise from structuring agent behavior around an explicit cognitive loop with external memory and a lightweight controller.

To support reproducibility and hands-on exploration, a live demonstration of the travel planner scenario described in Section 4 is available at https://scl-travel-planner.streamlit.app/.

## 5 Conclusion

The right measure of an AI agent is not how well it performs but whether its decisions are accountable—traceable to identifiable structural components, grounded in warranted evidence, and transparent enough to support attribution of responsibility when things go wrong. Performance and accountability are not the same axis. An agent that produces correct outputs through an opaque process provides no basis for institutional trust, error correction, or legitimate delegation. The core thesis of this paper is that accountable AI agency requires architectural design, not just better models.

### 5.1 What the Architecture Establishes

SCL demonstrates that separating cognition, control, action, and memory into structurally distinct modules produces behavior that is both more reliable and more inspectable. Across 360 evaluation episodes, task success averaged 86.3 percent against 70.5–76.8 percent for prompt-based baselines using the same model and tools. The gains are modest, but their source is identifiable: external memory prevents state loss, guarded execution prevents premature action, and the audit trail localizes errors to specific loop phases. Ablations confirm that memory and control each contribute independently. The significance of these results lies not primarily in the performance gap but in what the architecture makes possible: when an episode fails, the trace shows exactly where and why. That property—structural localizability of error—is the minimum condition for accountable agency, and it is precisely what prompt-centric architectures cannot provide.

### 5.2 Limitations and Subsequent Extensions

The evaluation is intentionally constrained. A single base model and fixed tool set were used to isolate architectural effects; generality across model families and task domains requires further validation. The loop as presented also does not address two accountability gaps that SCL's own design makes visible. First, human judgment is external to the loop: the architecture produces a trace, but provides no structural guarantee that human oversight is integrated into the decision process rather than appended after it. Second, retrieved evidence enters reasoning as context text without a principled mechanism for distinguishing warranted evidence from parametric inference—meaning that absence of information can be silently filled by hallucination. Both gaps reflect the natural boundary of what a behavioral loop alone can achieve. Subsequent research has addressed each in turn: context-aware Human-in-the-Loop embeds human judgment as a first-class cognitive event with cognitive state freezing and virtual rejection cycles; Pool-Gated Retrieval enforces a Horizon-Warrant-Commitment admission structure that registers confirmed absences as first-class facts. The Horizon-Warrant-Commitment framework then provides the philosophical foundation unifying both, establishing the structural conditions under which any epistemic input can be legitimately incorporated into an accountable agent. Together, these three lines of work extend SCL toward a more complete accountability architecture.

A further line of development revisits the loop itself (Section 3.10). By compiling instructions into programs that are validated before execution, it relocates rule compliance from model interpretation into deterministic structure and declines to execute whenever the meaning of an instruction does not close unambiguously. Its systematic evaluation remains an open task.

### 5.3 Closing Remark

The field has invested heavily in making agents more capable. This paper argues for a complementary and prior commitment: making agents accountable. Capability without accountability is an agent that may do the right thing without there being any principled reason why—and that cannot provide the institutional grounding that real-world deployment requires. SCL, Pool-Gated Retrieval, and the Horizon-Warrant-Commitment framework together constitute a research program in which accountability is the design objective, not a property to be inferred from performance after the fact. The architecture presented here is the foundation of that program. Its current direction, from an interpreted loop to compiled and validated programs, carries the same commitment from the level of action to the level of meaning. SCL is, in this sense, an accountability structure: it does not rest its accountability on the agreement of agents or on the imitation of human minds, but institutionalizes cognition so that every authoritative step can be attributed, examined, and corrected. What an agent is entitled to conclude—and the structural conditions under which that entitlement is conferred—is the question that accountable AI must answer.